# A Framework for Assurance of Medication Safety using Machine Learning

Yan Jia[1]   Tom Lawton[2]   John McDermid[1]   Eric Rojas[3]   Ibrahim Habli[1]

*Abstract*— **Medication errors continue to be the leading cause of avoidable patient harm in hospitals. This paper sets out a framework to assure medication safety that combines machine learning and safety engineering methods. It uses safety analysis to proactively identify potential causes of medication error, based on expert opinion. As healthcare is now data rich, it is possible to augment safety analysis with machine learning to discover actual causes of medication error from the data, and to identify where they deviate from what was predicted in the safety analysis. Combining these two views has the potential to enable the risk of medication errors to be managed proactively and dynamically. We apply the framework to a case study involving thoracic surgery, e.g. oesophagectomy, where errors in giving beta-blockers can be critical to control atrial fibrillation. This case study combines a HAZOP-based safety analysis method known as SHARD with Bayesian network structure learning and process mining to produce the analysis results, showing the potential of the framework for ensuring patient safety, and for transforming the way that safety is managed in complex healthcare environments.**

1. INTRODUCTION

Safety analysis is both predictive and reactive. It aims to identify hazards, hazard causes and mitigations, and associated risks before a system is deployed. The system is then monitored during its deployment to manage risks. When systems are used in highly controlled environments, built using components with a long service history, these predictions can be accurate. However, as systems become more complex, especially in adaptive socio-technical contexts, it is more difficult to make credible predictions of safety. Indeed, recent analyses of quantitative risk assessment (QRA) by Rae et al [1] show that analyses even of technical systems are rarely accurate, and the paper systematically identifies causes of deviations between prediction and reality in order to propose ways of improving QRA (and hence safety management). Other research, e.g. that of Hollnagel [2], employs the distinction between work-as-imagined and work-as-done, to acknowledge the gap between what is predicted and how the system actually operates and thus to explain the inaccuracies in safety analysis. They propose a number of solutions including using resilience engineering to manage safety more effectively [3, 4].

Healthcare as a complex social-technical domain covers a diverse set of activities that have to deal with its nonlinear, dynamic and unpredictable nature [5, 6]. This potentially increases the gap between work-as-imagined and work-as-done. The medication management process is very complex, including different phases, from prescribing, dispensing, administering to monitoring. For each phase, there are different clinicians involved, e.g. doctors, pharmacists, nurses, care givers, etc. Medication errors can arise in different phases and have many potential causes ranging from clinical factors, e.g. due to comorbidities, via technical factors, e.g. due to problems with Health IT (HIT) systems, such as electronic patient records (EPR), to human and organizational factors, e.g. under-staffing. They are variable and context-sensitive [7]. Understanding the variation and the significance of the causes of medication errors in different contexts is particularly important to support clinicians and healthcare organisations in anticipating, monitoring and responding to medication safety [8].

In addition, the recent emergence of big data in healthcare, including large linked data from EPR as well as streams of real-time health data collected by personal wearable devices [9], gives the opportunity

[1] Department of Computer Science, University of York, York, UK. Correspondence: Yan Jia <yan.jia@york.ac.uk>.
[2] Bradford Royal Infirmary and Bradford Institute for Health Research, Bradford, UK
[3] Internal Medicine Department, School of Medicine. Pontificia Universidad Católica de Chile, Santiago, Chile

to use data analysis methods, including Machine Learning (ML), to better understand the nature of medication errors, their causes and their impact in context.

Our contribution in this paper is twofold. First, it presents a new framework that provides a practical means of reducing the gap between the work-as-imagined and the work-as-done, from a safety perspective. This can be thought of as actualising some of Hollnagel's ideas, but a unique contribution is to do so via the use of ML and other analytical techniques on data obtained from the real world, which we call the 'work-as-observed'. Second, it shows that the framework can be applied to management of medication safety, giving clinically meaningful results. This enables medication safety to be managed effectively and dynamically, using the results of ML.

The framework is illustrated and validated on healthcare applications although we believe that the approach could have wider usage [10]. The use case we present considers the complex setting of intensive care units (ICU) where patients may be taking multiple medications due to comorbidities, and the post-operative care is perhaps the most difficult to manage, especially when the treatment is time-critical. This work is extremely important as patients in ICUs are at high risk and medication errors can be life-threatening [11].

The rest of the paper is structured as follows. Section 2 discusses related work, including the use of ML in medication safety. Section 3 presents our framework, including explaining the role of a safety case in recording and presenting the results of analysis of work-as-imagined and work-as-observed. Section 4 presents our case study, the management of atrial fibrillation in post-operative care in an ICU following thoracic surgery. The case study uses a safety analysis method known as Software Hazard Analysis and Resolution in Design (SHARD) [12] for hazard and risk analysis. Then it employs ML, specifically Bayesian network (BN) structure learning, and process mining to analyse hospital data and provide evidence for an explicit safety argument. The potential for wider use of the framework and some potential issues with the use of ML are addressed in Section 5. Section 6 presents conclusions.

## 2. RELATED WORK

This section considers medication safety, then discusses the use of ML both in clinical decision support systems and in support of safety analysis.

### 2.1. MEDICATION SAFETY

According to the World Health Organisation there is now overwhelming evidence that significant numbers of patients are harmed by their healthcare, resulting in permanent injury, increased length of stay in hospitals, and death [13]. We have learnt over the last decade that patient harm occurs because the healthcare system today is so complex that the successful treatment for each patient depends on a range of factors, such as the competence of health-care providers, staff rosters and the effectiveness of diagnostic systems. As so many types of healthcare providers are involved, it is essential to ensure that the healthcare system as a whole is adaptive so as to enable timely and effective interaction and shared understanding of risks and uncertainties by all the healthcare professionals.

Improving medication safety has become a priority in healthcare organisations to improve patient safety [14] and it has a critical impact on patient outcomes. As medication errors continue to be a leading cause of avoidable harm in hospitals [15, 16], both regulatory agencies and research communities have made efforts to improve medication safety. One of the recommendations to reduce medication errors is to use the 'five rights': the right patient, the right drug, the right dose, the right route, and the right time [17]. These can be seen as overarching goals to achieve medication safety. However, they are broadly stated and are not supported by procedural guidance on how to achieve them [18]. Other literature on medication error, e.g. [19, 20], presents a statistical analysis of medication errors across the whole process from prescription to administration. The analyses typically show that a high proportion of prescriptions in hospitals are subject to some form of error although, of course, the majority are corrected prior to administering the drugs.

As well as statistical analyses, there is work intended to establish practical and proactive means for identifying and detailing the underlying causes of errors and finding potential controls for those errors, e.g. [21, 22]. The Safety Assurance of Intravenous Medication Management Systems (SAM) project

[23, 24] is focusing on using technology to automate cross-checks to reduce certain classes of error. This is an innovative approach to the issue and is notable for considering the acceptability of the technology to patients and clinicians.

The intent of our work is to identify practical ways of narrowing the gap between work-as-imagined and work-as-done, using ML to understand the differences, hence improving medication safety. We view this as complementary to, and different from, the approaches identified above.

### 2.2. ML IN CLINICAL DECISION SUPPORT SYSTEMS

As well as the effort in the clinical safety community, the artificial intelligence (AI) research community is investigating the use of ML techniques in healthcare to improve patient safety. ML-based clinical decision support systems are being developed to provide guidance on the safe prescription of medicines, guideline adherence, diagnostic decision support and prognostic scoring [25]. ML is often used to support diagnostic decision-making, specifically in clinical domains such as radiology, using algorithms that learn from training data to classify images [26-28]. Significant examples of such usage of ML include the identification of malignant lesions and cancers from skin photographs [29, 30], analysis of echocardiograms to detect heart problems, e.g. hypertrophic cardiomyopathy and pulmonary artery hypertension [31], and prediction of sight-threatening diseases from eye scans using optical coherence tomography [32].

Outside of diagnostic support ML systems are being developed to provide other kinds of decision support, such as treatment recommendations [33, 34]. For example, the AI Clinician uses reinforcement learning to devise effective treatment strategies for sepsis, which is the third leading cause of death in hospitals [35, 36]. Komorowski et al [34] argue that the value of the AI Clinician's recommended treatment is on average reliably higher than human doctors, although this is not based on operational deployment of the tool. Other work on decision support provides risk predictions where many complex and interacting factors have to be taken into consideration. For example, one project has used BN to predict the risk of developing coronary heart disease [37], based on life-style data collected over many years. Another project has used ML to predict the risk of suicide attempts [38], again based on complex data.

Research has also been focused on automatic triage for patients or prioritising individual access to clinical services by screening referrals. Babylon Health has an ambitious mission: to put an accessible and affordable health service in the hands of every person on earth. They have a growing range of services [39], including an on-line triage tool which analyses data provided by patients to advise them on a course of action, e.g. to consult a physician or to visit a pharmacist. The tool uses a range of ML technologies, e.g. recurrent neural networks for processing and analysing the text input by the user. It also employs an extensive medical knowledge base to assist in interpreting the information on symptoms provided by the user [40]. Whilst Babylon see their technology as helping address some of the problems of clinician shortages, their work is not without its critics [41].

Our work, in contrast to all of these approaches, is using ML as part of the framework to bridge the gap between the work-as-imagined and the work-as-done, and thus to improve the assurance of medication safety, rather than supporting diagnosis or other medical activities directly.

### 2.3. ML IN SAFETY ASSURANCE

As outlined above ML techniques have been applied to different medical applications with an aim of improving safety in healthcare, e.g. diagnosis and treatment practices. It has not been directly used to support the understanding of risk and the safety assurance process itself. Here we present a few examples of work that has used ML to understand or support safety directly. We note that, in order to find such examples, we need to look outside healthcare.

An example from the Oil and Gas industry [42] uses deep neural networks on risk assessment for (unintended) movements of the platform, which might ultimately lead to damage to the wellhead. The authors are cautious about their findings and note that care needs to be taken in selecting models to support safety-related decision making; we return to this point in section 5. Further, there have been accidents with Unmanned Air Systems (UAS), e.g. Watchkeeper, where the accident causation was very different to that predicted in the initial safety analysis [43]. Work is under way using ML to identify the causal factors that contributed to the accidents, including differences between the UAS' behaviour and the operators' perception of what was happening [44].

The use of ML in support of safety assurance is novel but there is an initial awareness of its potential; our framework is intended to help realise this potential.

3. PROPOSED FRAMEWORK

Before introducing our framework, we give a brief overview of traditional and modern approaches to safety management, to provide the context and some of the building blocks for our framework.

3.1. TRADITIONAL AND MODERN APPROACHES TO SAFETY MANAGEMENT

Safety management has traditionally focused on failure – implicitly assuming that effective system design and compliance with good operating procedures will be safe. It also assumes that it is possible to analyse potential failures and to design mechanisms or procedures to control these failures and thus to assure safety. Hollnagel characterised this as 'Safety-I' and states that the purpose of safety management in Safety-I is to keep the number of accidents and incidents as low as possible by reacting to unacceptable events, due to system malfunction or human fallibility [45]. This does not properly cater even for the current socio-technical systems and will become even less effective as they become more complicated and hence intractable. Hollnagel emphasises that safety management should therefore move from ensuring that 'as few things as possible go wrong' to ensuring that 'as many things as possible go right' which he characterises as 'Safety-II' [45]. He states that it is necessary to adopt Safety-II, which essentially means taking a resilience engineering perspective.

In part this initiative builds on the distinction between 'work-as-imagined' and 'work-as-done' [46], with the terms used as follows:

- Work-as-imagined – how work is thought of either before it takes place when it is being planned or after it has taken place, when the consequences are being evaluated;
- Work-as-done – how work is actually carried out, where and when it happens.

This distinction is especially valuable where the system is not closed, as descriptions of the work have to assume some nominal conditions, e.g. good clinical practice, which may not be the case all the time. For example, in healthcare where unexpected events can happen, it is impossible for humans to predict all the conditions that can arise. Hence, performance adjustments are necessary to enable the work to be carried out successfully, resulting in differences between work-as-imagined and work-as-done [47]. Therefore, work-as-imagined and work-as-done make sense in their respective contexts, but sometimes they may not be aligned. The idea of realignment between these two acknowledges, on the one hand, that neither work-as-imagined nor work-as-done are an absolute reference (for reasons, please see [46]), and on the other hand, that the gap between them should be reduced as much as possible. Our framework builds on this distinction, adding the concept of work-as-observed.

Other notable innovations include Leveson's STAMP/STPA [48] which adopts a system theoretic view of safety management and addresses some key aspects of socio-technical systems. There have been applications of STAMP/STPA in healthcare, e.g. for analysing patient safety incidents including medication error [49]. However, this does not provide an effective basis for analysing the distinction between 'work-as-imagined' and 'work-as-done', and we do not build on it further here.

3.2. THE FRAMEWORK

The motivation for us to develop this framework for medication safety assurance is to provide a solution to realigning work-as-done and work-as-imagined, using ML and safety cases. We see this as a way of implementing Safety-II, which can be seen more as a broadly stated concept than as a practical methodology.

The framework is presented in Figure 1. It does not include specific methods; the methods should be chosen for specific applications of the framework, see section 4 for an illustration. What we have developed here is a combination of Safety-I and Safety-II (we believe these two views are complementary and not alternatives). It uses traditional safety analysis techniques to analyse the normative model to generate the failure model to represent the work-as-imagined. This is the Safety-I point of view. Then the framework uses ML to learn from the data which is generated from the real world, which we call work-as-observed here. This is the Safety-II point of view. By work-as-observed, we mean that the data which is generated from the real-world can be analysed to give insights into what is actually happening and thus to help understand the gap between the work-as-imagined and work-as-done.

The framework comprises four elements:

- Real-world, which is intended to be broader than Hollnagel's 'work-as-done', includes human activities, and technological systems as well as the hospital environment;
- Work-as-imagined, which we use in a similar sense to Hollnagel [46] but explicitly including normative models of work and safety analysis of those models;
- Work-as-observed, which is intended to provide the model of the real world through ML techniques on available healthcare data sets;
- Safety case [50], which we use to reflect the analysis in the work-as-imagined, but also incorporating the understanding gained from analysis in the work-as-observed to enable the different viewpoints to be realigned.

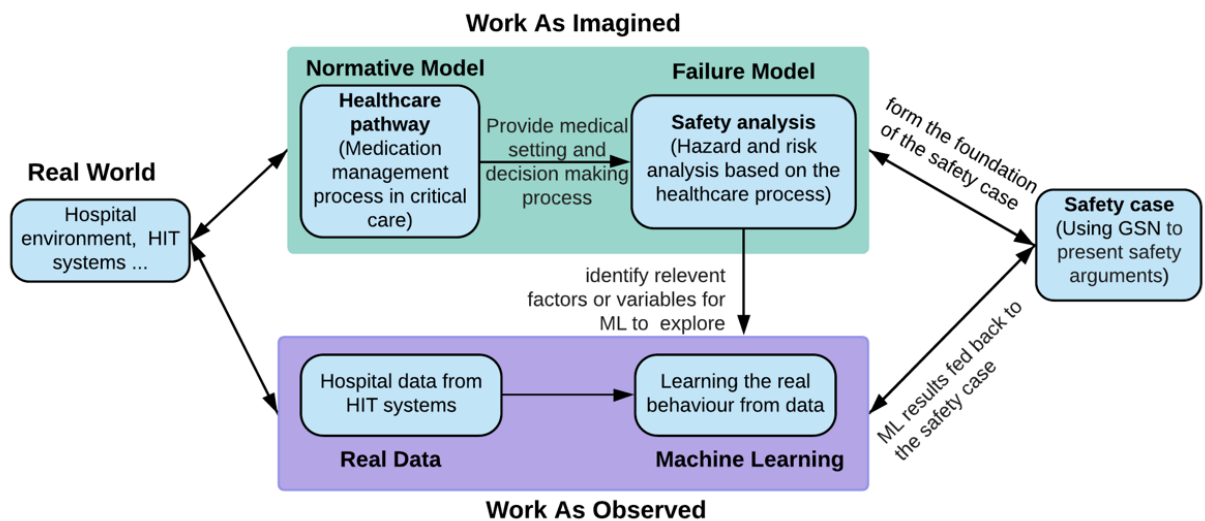

Fig. 1. Framework for Modern Safety Management

More specifically, the framework uses healthcare pathways as the normative model of the healthcare processes which serves as a basis for safety analysis to provide the failure model. Safety analysis has two components. First, identifying hazards associated with this healthcare system, in its context of use. Second, determining the potential causes and effects of the hazards, together with severity (degree of harm) and likelihood of the hazard, to estimate risk. Together these form the work-as-imagined.

Traditionally, especially during system design, safety analyses represent expert judgment of how the system will work, what might cause the hazards, and how they are controlled. In the healthcare setting, this would reflect the analysis of the normative models employing both systematic safety analysis methods and domain knowledge. Our framework is 'agnostic' with respect to the safety analysis methods that are used, however we illustrate the framework in the case study using a flow-based analysis method - SHARD, as this is effective in assessment of process models such as clinical pathways and helps to identify the factors that can contribute to medication errors.

Work-as-observed is what we can infer from available data. With the growing volume of EPR and other hospital data available, it is possible to learn more from data, e.g. to identify patterns of how clinicians carry out their work. Having a better understanding of the work-as-observed is particularly helpful to understand what is actually happening (real world) in healthcare, in the context of the complex processes and the variability between patients. Using ML on data from the hospitals post-deployment, e.g. from EPR systems, can help us to explore the dependencies between the factors identified in the safety analysis or even identify other factors that we did not consider pre-deployment. Using variables to represent the hazards, causes of hazards, and effects of hazards from the safety analysis, it is possible to identify the dependencies in the data to validate (or find limitations in) the safety analysis. As with safety analysis our framework is 'agnostic' to particular ML methods, but the example here uses BN structure learning and process mining as they are effective in identifying dependencies and they can reveal fine structure in complex data sets. What is worth mentioning here is that the value of the analysis will all depend on how good the data is.

The results of the ML analysis will reflect the work-as-observed and will be reconciled with analysis of the work-as-imagined, to provide the basis for the safety case. The safety case is an argument (in the sense of a rationale) supported by evidence to show that the (socio-technical) system is safe [51]. As neither work-as-imagined nor work-as-done should be viewed as an absolute reference, the safety case can be used as a medium to realign work-as-imagined and work-as-done using ML to continuously learn from real hospital data, i.e. work-as-observed. This will help the healthcare manager or other clinical decision-makers, who are at the 'blunt end' to see the information more precisely. In turn, this could also help the decision-maker to understand when it is desirable to change work-as-imagined, e.g. improving the normative model or data collection methods from EPR if there is value in doing so. Additionally, by improving the normative model, establishing new care pathways, improving data collection methods, or introducing other interventions, we intend that this can influence the real-world and improve patient safety. The feedback paths in Fig. 1 from the safety case to the work-as-imagined and the work-as-observed are intended to indicate these improvements.

It is noteworthy that the gaps between work-as-done and work-as-imagined can never be completely eliminated in a complex socio-technical setting, as the work-as-done is dynamic and evolving. But ML can be used in the work-as-observed to help identify these gaps, and the safety case can then help to visualise them. Further, this could give the basis for determining what sorts of intervention are needed to reduce the gap. Therefore, the feedback loops need to operate continuously to reflect and control the gaps.

## 4. CASE STUDY

The case study is intended to show how to implement this framework and to evaluate it. It focuses on medication management for patients taking beta-blockers (BBs) before surgery involving the thorax. An overview of the clinical setting is given in Section 4.1, followed, in Sections 4.2 to 4.4, by an instantiation of the framework to show how it has been applied to the case study.

### 4.1. CLINICAL SETTING

Patients undergoing thoracic surgery are at risk of disturbances of heart rhythm, typically atrial fibrillation (AF) in post-operative care after opening the chest [52-54]. There is debate about whether or not all patients should receive BBs following surgery but, as a minimum, treatment should continue to give BBs following thoracic surgery to reduce the risk of AF for those who were receiving BBs before surgery [55, 56].

Oesophagectomy is a thoracic operation whereby the oesophagus (food pipe) is removed, usually to treat oesophageal cancer. We use oesophagectomy as a concrete example of thoracic surgery in this case to illustrate the framework. As oesophagectomy prevents the patient from taking food or drugs orally, especially in the first week after surgery, the challenge is how to give the right form of BBs when the patient is unable to swallow. If BBs are not continued post-operatively, then there is an increased risk of developing AF which can lead to strokes that may be fatal – one analysis gives an odds ratio of death from AF of 1.5 for men and 1.9 for women [57].

There are a number of published guidelines on post-operative care for oesophagectomy. The pathway in Fig. 2 has been synthesised from a number of publications [58-60] and focuses on the delivery of nutrition and medication. The pathway shows the differences between the presence or absence of a feeding tube (FT). Patients may be fitted with a FT during the operation and this also can be used for some forms of medicine. If there is no FT, medication has to be given by intravenous (IV) injection or infusion. This increases the complexity of giving BBs post-operatively, as commonly used BBs, e.g. bisoprolol, do not have an IV form. Also, the IV form of BBs, e.g. metoprolol and atenolol, may be less familiar to clinicians, and may not be immediately available on the ward. Further, the calculation of equivalent doses makes mapping between oral and IV form of BBs error-prone.

A further complicating factor is that patients may have other medications, e.g. painkillers given epidurally, and there can be an adverse interaction with BBs leading to a potentially dangerous reduction in blood pressure (BP). Overall, there are many potential difficulties in managing delivery of BBs in post-operative care. This makes it a rich case study for our framework.

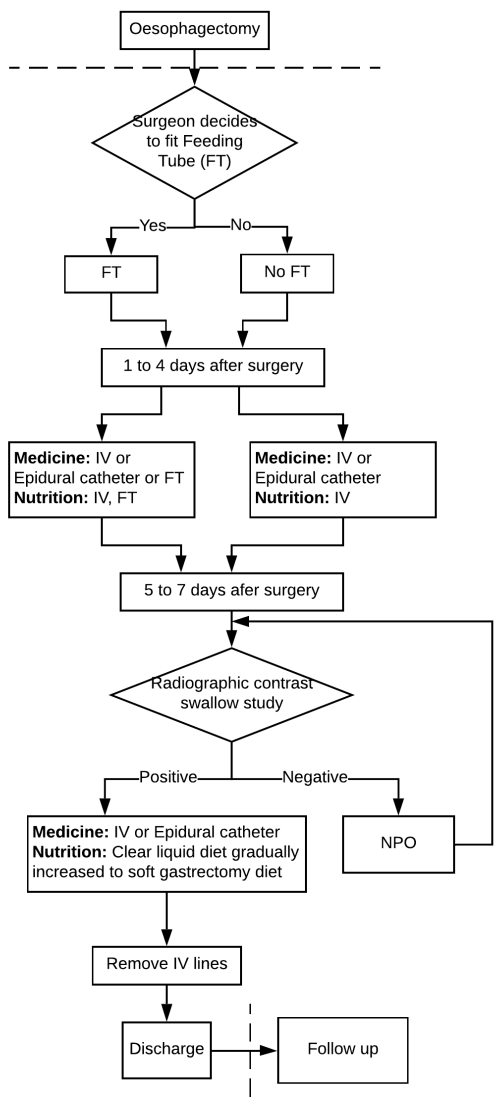

Fig. 2. Pathway for Nutrition and Medication following oesophagectomy

### 4.2. INSTANTIATION OF WORK-AS-IMAGINED

The work-as-imagined contains normative models and the safety analysis of these models. The normative model comprises a decision-making model related to delivery of medication in post-operative care following an oesophagectomy.

The normative model gives the medical setting for conducting the safety analysis. Section 4.2.2 presents the associated safety analysis recorded as the failure model. It uses SHARD method, to identify hazards, their potential causes and controls.

#### 4.2.1. NORMATIVE MODEL

The simplified decision-making model is shown in Fig. 3, which identifies the roles of the different professionals involved in this process. The main stakeholders in this process are the doctor, the pharmacist and the nurse with responsibilities for the prescribing, dispensing and administering phases, respectively. The surgeons are also included as they normally determine whether or not an FT is fitted, which serves as an important context for the medication decision making. In practice, the situation is more complex, e.g. as more than one nurse will be involved in administering medications and sometimes because of understaffing, the nurses might not be able to administer the medicine as expected, but Fig. 3 is intended to map the clinical roles rather than the work of particular individuals.

In Fig. 3 the 'administration' outcome represents the success of the overall medication management process. Fig. 3 also shows three types of failure: fail to prescribe, fail to dispense and fail to administer – the outputs of decisions C, D and F respectively. These are cases where the BB will not be administered, which might increase the risk of AF. There are also potential controls for these failures. In general, these involve the other professionals (nurse or pharmacist) referring back to the doctor for resolution, e.g. if the medication is detected to be of an inappropriate form to administer. In addition, this decision-making model shows the complicating factors, e.g. the presence of a feeding tube and the need to calculate equivalent doses, that we mentioned in Section 4.1.

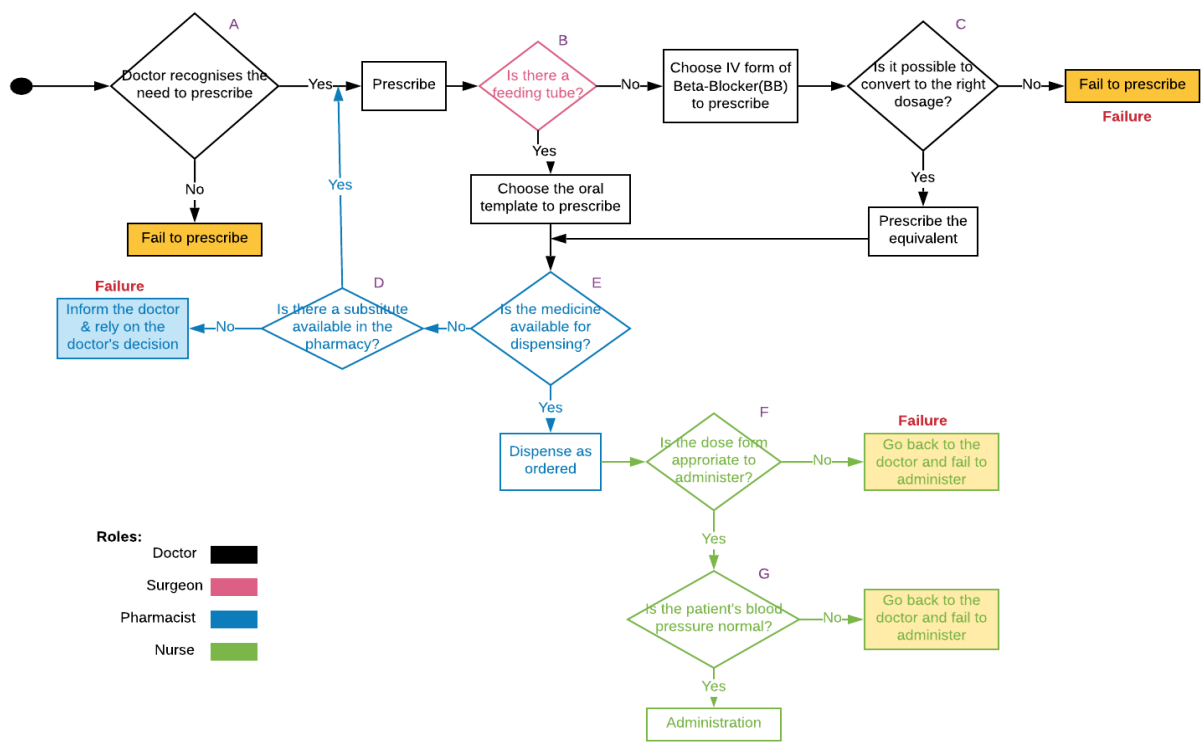

Fig. 3. Decision-making flowchart for prescription and administration of medication

### 4.2.2. FAILURE MODEL

In safety management it is common to organise risk analysis and control around the notion of a hazard – a situation which, if not controlled, could lead to harm [48]. This is quite a general definition, and it needs to be interpreted in the context of a particular system or situation [61].

Once hazards have been identified the system or situation is analysed to determine potential causes of the hazards. In some industries it is common to quantify the likelihood of these causes and to use these figures to prioritise the introduction of risk controls (means of preventing the causes of hazards or reducing the impact of hazards if they do arise). Once controls have been identified then the risk can be re-evaluated.

Where it is difficult to quantify the likelihood of hazard causes, the analysis is typically qualitative, and judgment is needed on the choice of controls to manage risk cost-effectively. The discussion below is largely qualitative, but ML offers credible opportunities to quantify risks. Such analysis can be used to guide risk management including the introduction of additional risk controls (this is a key part of the feedback path from work-as-imagined to the real-world).

Hazard and safety analysis of computer-controlled systems often use variants of Hazard and Operability Studies (HAZOPS) [62] from the chemical industry. HAZOPS are based on flows (of chemicals, etc.) through process plant. The variants used for computer-based systems consider information flows through systems. SHARD is suitable for identifying both hazards and causes of hazards, as it focuses on deviations from intent that could be hazardous. It provides a structured approach to the identification of deviations by systematically applying the guidewords (*omission*, *commission*, *early*, *late* and *incorrect*) to each flow. In this context, commission means doing something that was not intended.

Having identified hazards, it is necessary to determine potential causes of hazards as this gives a basis for defining controls and to reduce risks. We also perform this using SHARD with the results presented in a table with the following columns: the *guideword* applied to the flow of information, the interpretation of that guideword (*deviation*), *possible causes* of the deviation (either local failures or incorrect inputs from earlier in the process), ways of *detecting* the deviation and *protecting* against it, and finally the *potential effects*.

Identifying hazards is often carried out by experts who determine specific situations which could give rise to harm, prompted by the guidewords. In the context considered in this paper, the hazards identified from the use of the SHARD guidewords, as confirmed with the judgement of medical experts, are as follows (also see column 2 of Table I):

- Omission – failure to administer BB to patients who took BB pre-operation (Hazard 1);
- Commission – unnecessary BB administered (Hazard 2);
- Incorrect – patient receives wrong BB (e.g. contra-indication with other medications or comorbidities) (Hazard 3);
- Incorrect – underdosage of BB (e.g. not all prescribed doses administered) (Hazard 4);
- Incorrect – overdosage of BB (e.g. repeated administration of doses) (Hazard 5).

However, it is necessary to note that there may be an appropriate omission in order to reduce the possibility of a different harm - worsening hypotension despite the potential increased risk of AF. Timing issues (*early* and *late*) are not considered as hazard categories here, as any harmful effects would be 'caught' by the under and over cases of the incorrect hazard category.

The clinical outcomes from these hazards could vary significantly. AF is most likely to be caused by omission (a failure to administer BBs). The effect of 'incorrect' depends on the medication administered; it might just cause dizziness, although the worst-case outcomes might be more severe.

Here we apply SHARD to the decision-making model in Fig. 3, with the intent of finding causes of the hazards and ways in which the decision process can be improved to reduce the likelihood of error, and the results are presented in Table 1. Using SHARD, it is common to work through the whole system description, starting at the end and working backwards. So, we illustrate the SHARD analysis (in tabular form) starting at the end of the decision-making model, i.e. the administration. In addition, we also include technical and organisational factors that are not explicit in the decision-making model (the entries in blue). The rows correspond to the hazards identified earlier (column 2). The entries in the *detection/protection* column include use of EPR to make recommendations on order entry but are heavily dependent on the medical staff.

As mentioned above, risk assessment in this context is essentially qualitative. The most likely, or credible, causes of hazards can be identified typically using expert judgment. In this case understaffing of wards might be the most likely cause of failure to administer BBs to patients following a thoracic operation. This information can then be used to prioritise the introduction of new controls. However, even this qualitative approach can be difficult in a medical setting because of the many shaping factors such as the patient's general health, comorbidities, etc. This is a further motivation for using ML to complement the safety analysis. We show in Section 4.3 that hazard 1 – failure to administer BB to patients following a thoracic operation – would cause a 11% increase in the likelihood of AF post-operation, a result which no qualitative analysis could produce.

TABLE I. SHARD RESULTS OF THE DECISION-MAKING MODEL

| Guideword | Deviation | Possible Causes (Labels correspond to decisions in Fig. 3) | Potential Detection/Protection | Potential Effects |
|-----------|-----------|------------------------------------------------------------|--------------------------------|-------------------|
|           |           |                                                            |                                |                   |
| Omission | No BB administered (Hazard 1) | ① G. Patient is suffering hypotension (may be due to epidural) and nurses decide not to administer<br><br>② B, F. Wrong form of BB prescribed or dispensed for available route, so nurses do not administer<br><br>③ A, C. No BB prescribed or dispensed<br><br>④ Understaffing of wards leads to doses being missed (organisational factor)<br><br>⑤ Complete failure of IV device or infusion pump (technical factor) | 1. Clinicians should check BP and medications using drug chart on a daily basis<br><br>2. Pharmacist should review prescription with clinician if suitable medication unavailable<br><br>3. Nurse should identify the wrong form and query with clinician | AF |
| Commission | Unnecessary BB administered (Hazard 2) | A. Unnecessary BB prescribed<br>E. Unnecessary BB dispensed<br>Busy ward leads to administering medicine for wrong patient (organisational factor) | 1. Pharmacist should review the prescriptions<br><br>2. Nurses should check the prescriptions before administering | Adverse interactions with other medication or comorbidities |
| Incorrect | Wrong BB administered (Hazard 3) | D. Incorrect substitution | 1. Pharmacist should review the prescriptions | Adverse interactions with other medication or comorbidities |
| Incorrect | Under dosage* (Hazard 4) | C. Incorrect dose calculation<br>G. Patient is suffering hypotension (may be due to epidural) and nurses decide not to administer<br>Understaffing on wards leads to some doses being missed (organisational factor)<br>Inappropriate recommendation from EPR (technical factor)<br>Rate error of IV device or infusion pump (technical factor) | 1. Order entry from the EPR might help the clinician to prescribe correct dosage<br><br>2. Pharmacist might pick up the error<br><br>3. Nurses might pick up the error | AF |
| Incorrect | Over dosage* (Hazard 5) | C. Incorrect dose calculation<br>A. Doctor might prescribe both forms of BB to let the nurse choose the suitable one and both doses are given to the patient.<br>Inappropriate recommendation from EPR (technical factor)<br>Rate error of IV device or infusion pump (technical factor) | 1. Order entry from the EPR might help the clinician to prescribe correct dosage<br><br>2. Pharmacist might pick up the error<br><br>3. Nurses might pick up the error | Hypotension |

*Assumes correct medication.

### 4.3. INSTANTIATION OF THE WORK-AS-OBSERVED

The work-as-observed uses ML to explore the 'ground truth' based on hospital data, i.e. to reflect the real-world. Here we use BN structure learning and process mining to understand the relationships between the different factors we extracted from safety analysis in Section 4.2 that might compromise medication safety. This is used to reason about the accuracy and soundness of the safety analysis.

The MIMIC III database [63] is a widely available online clinical data set, collected in ICU in the USA. We use it to provide the 'real data' for the work-as-observed. For brevity, we use Hazard 1 – no BB administered – as an illustration in this section. It focuses on the ML process and starts with a discussion of the role of data analysis in the framework and culminates in an analysis of the importance of administering BBs in controlling AF, based on the learnt models.

### 4.3.1. THE ROLE OF DATA ANALYSIS

Following our safety analysis, we have identified that factors such as the presence of hypotension (maybe due to epidural) can influence the decision to administer BBs. To understand whether or not these factors really are significant requires analysis of real data from hospitals. Whilst this could be seen as simply confirming expert opinion, experts' views can vary, so being able to base the results on extensive datasets helps to resolve inconsistent opinions. That is, the safety analysis has given us a hypothetical view of the causes and effects (the work-as-imagined) reflecting Safety-I. However, the actual risk still needs be evaluated and validated based on real clinical data (the work-as-observed) supporting Safety-II.

Given the clinical context, there are three primary variables for us to consider, *Pre_beta*, *Surgery*, *Post_beta* (see Table II). The reason why we categorise *Surgery* based on whether it involves the thorax or not is that, 1) any major thoracic surgery (not only oesophagectomy) carries the risk of AF in post-operative care especially for patients taking BBs before the surgery [54]; 2) in order to get more data to carry out ML, it is useful to consider all thoracic surgery rather than just oesophagectomy. Thus, the data analysis should identify these patients, along with the potential causes and effects of medication error, i.e. the 3$^{rd}$ and 5$^{th}$ columns in the 1$^{st}$ row in Table I, which means that we should consider: *hypotension*, *epidural*, *busy ward*, *understaffing of wards*, *failure or error rate of IV device or infusion pump*, from the *causes* column and *AF* from the *effect* column.

In this paper, we only focus on a subset of the potential causes of Hazard 1 – *Hypotension* and *Epidural*, because of the scope of the MIMIC III dataset. We expect to expand the analysis to include both organisational and technical factors using further data from Braford Teaching Hospital in England. This is discussed under future work.

### 4.3.2. DATA PREPARATION

All the SQL queries used to extract the data for this paper are available online at https://github.com/Yanjiayork/papers. Six of the variables identified above can be found in the MIMIC III data set and are described in Table II.

TABLE II. VARIABLES EXTRACTED FROM MIMIC III DATASET

| Variables | Variable Code | Variable Values |
|---|---|---|
| Surgery | Surgery | Value = 0, 'not thoracic'<br>Value = 1, 'might be thoracic'<br>Value = 2, 'definitely thoracic' |
| Receiving BB before surgery | Pre_beta | Value = 0, 'not receiving BB'<br>Value = 1, 'receiving BB' |
| Receiving BB after surgery | Post_beta | Value = 0, 'not receiving BB'<br>Value = 1, 'receiving BB' |
| Hypotension | Hypotension | Value = 0, 'no Hypotension'<br>Value = 1, 'has Hypotension' |
| Epidural catheter placed | Epidural | Value = 0, 'no Epidural'<br>Value = 1, 'has Epidural |
| Having AF during the encounter | AF | Value = 0, 'no AF'<br>Value = 1, 'has AF' |

We used the Current Procedural Terminology (CPT) codes [64] in MIMIC III to determine whether patients definitely had thoracic surgery, definitely did not, or possibly did (where there are alternative ways of doing the operation). For example, CPT code 43415 is defined as 'suture of an oesophageal wound or injury; transthoracic or transabdominal approach' which clearly can be conducted via the chest or the abdomen, hence we give it the value 1. On the other hand, CPT code 31760 is 'tracheoplasty; intrathoracic' which is definitely thoracic surgery, hence is given the value 2, and, of course, excision procedures on the oesophagus (oesophagectomy) are all given value 2.

The MIMIC III records are time-stamped, and the records are analysed to identify patients taking BBs at any time before an operation whilst in hospital (*Pre_beta*), or within 24 hours after surgery (*Post_beta*), as this is the most critical time. The value for *Hypotension* is based on the first reading after 6am on the day following surgery (less than 100mmHg is viewed as Hypotension and given the value 1). This time is chosen as it is when patients would normally have their BBs, so this BP reading is the one that is most likely to affect the nurse's decision about whether or not to administer the BB (this is entry ① for hazard 1 in Table I). Information about *AF* is inferred from the International Classification of Diseases, Ninth Revision (ICD 9) code, beginning with 427.

After the data preparation, 7,202 encounters were identified as relevant to this study, and had associated 'flags' indicating whether or not the patients suffered from AF, etc.

### 4.3.3. LEARNING BAYESIAN STRUCTURE FROM DATA

A BN is a directed graph of nodes and edges connecting those nodes. Each node represents a random variable, while the edge between the nodes represents probabilistic dependencies among the corresponding variables. Associated with each node is a conditional probability table, which specifies the probability of each node state given every combination of states of parent nodes. We use BNs as they have the potential to reveal a much finer structure by distinguishing between direct and indirect dependencies, by comparison with other statistical methods, such as logistic regression which focus on the most significant correlations.

BNs have a two-phase lifecycle. First, they are constructed, either by hand based on domain knowledge or by 'structure learning' from observational data [65]. In this case, the structure of a BN will be learned automatically using ML. To do this requires that we define a hypothesis space of possible structures for searching as well as a score function to measure each structure by a defined searching algorithm, such as greedy search [66]. In this paper, we use a greedy search-and-score methodology to learn the BN structure. BN structure learning provides a means to make sense of the complex correlations in clinical data that have hampered other approaches. Secondly, it is necessary to determine the probability distribution of each node in order to fully specify BNs.

The structure of the BN was learnt from the six variables shown in Table II using BDeu (Bayesian Dirichlet equivalence uniform) score [67] based on the data preparation described in section 4.3.2. BDeu is a widely-used scoring metric for learning BN structures for complete discrete data. Fig. 4 presents the results. Note that arrow directions in the structure learnt should not be interpreted as showing causality, only a statistical correlation.

The model in Fig. 4 generally reflects the safety analysis in Table I, but there is one point of interest. It shows no direct dependency between *Epidural* and *Hypotension*, despite the fact that it shows *Post_beta* has an individual direct dependency with both *Hypotension* and *Epidural*. This is a very interesting finding, as our safety analysis shows that *Hypotension* should be the reason for patients not receiving *Post_beta* rather than *Epidural*. *Epidural* might affect *Post_beta* because it has the potential to cause *Hypotension*, but itself should not influence whether or not to administer BB. So, we expect there to be a direct dependency between *Epidural* and *Hypotension*, especially when *Epidural* has a direct dependency with *Post_beta* as shown in fig. 4.

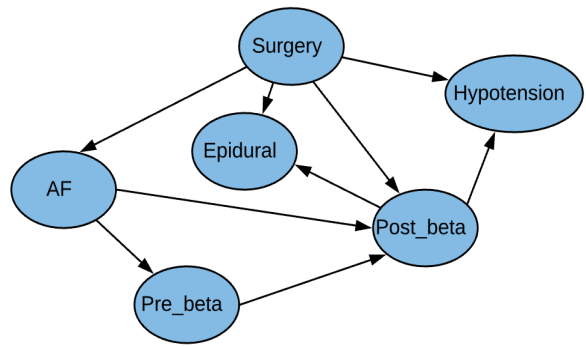

Fig. 4. Learnt Bayesian Network structure based on Safety Analysis

We initially thought that it might be to do with confounding factors, e.g. normal BP readings will be obtained if they are measured whilst vasopressor medications such as phenylephrine or nor-epinephrine are being given by infusion. However, even when we 'corrected' the value of the variable by considering *Hypotension* to be present, although the BP reading is normal, whilst such infusions were being given did not alter the learnt structure. On reflection, this might be best explained as follows.

*Epidural* can cause *Hypotension*, but it might just cause a slight drop of the BP, not sufficiently severe to count as *Hypotension* [68]. Also, the means of giving *Epidural*, e.g. infusion or bolus will influence the result. If *Epidural* is given by bolus, it is more likely to cause *Hypotension* [69]. In addition, different patients will react differently and often the patients can bring up their BP in a short time (then keep it up) with their own bodily mechanisms. As we chose one time to read BP, this might not be immediately after epidural. Thus, it is not surprising that there is no direct dependency between *Epidural* and *Hypotension*.

However, Fig. 4 might suggest a new pattern of how nurses carry out their work in the real world. If a nurse is aware that the patient has an epidural (and it should be quite obvious) a decision might be made not to give the BB, even when their BP is normal as they know the potential effect of epidural and BBs on BP. The combination of these two events is capable of causing a severe drop in BP in some situations [69]. Thus, a direct dependency is learnt between *Epidural* and *Post_beta*, even when there is no direct dependency between *Epidural* and *Hypotension*.

In order to explore whether this suggested pattern for epidural patients is valid or not, we decided to use process mining to do further investigation as this allows temporal relationships in the data to be exposed.

### 4.3.4. PROCESS MINING

A process is viewed as a series of actions or steps forming a progression from a defined or recognized 'start' to a defined or recognized 'finish'. In the context of our case study, the series of activities – prescribing, dispensing and administering a drug is an example of a process. Process mining is a data-oriented process analysis field that aims to discover real-world processes by extracting knowledge from available event logs [70]. An event log corresponds to all the activities executed during time by a specific resource or resources, e.g. doctors or nurses, in a given healthcare setting. The executed activities are stored in the EPR systems, being available for us to build the event log and execute multiple process mining methods and techniques. In our terminology, this is analysing the real data from an organisation's systems in the world-as-observed.

One of the main analyses available in process mining is the discovery analysis, which takes the event log as an input and generates, through specific algorithms, a process model [71]. A process model reflects each of the executed activities, with arcs connecting each activity to reflect their execution order. The process model also shows the number of executions for an activity or for an arc between two activities. Often, if process mining is used with all the available data and not to answer specific questions, it will produce a 'spaghetti' model with everything connected to everything else [72, 73]. Thus, it is desirable to focus the use of process mining onto specific questions.

For the discovery analysis in process mining, the question we focused on is whether or not the work pattern we suggested for epidural patients in 4.3.3 is the case. To answer this question, we extracted the subset of the data used for structure learning that contained the patients who had epidural. As some of the patients did not have time stamps for epidural recorded in MIMIC III dataset, this gave us 141 cases of epidural for process mining; the results of this discovery analysis are shown in Fig. 5 below. Fig. 5 provides a model with 6 activities each containing a number that indicates the total amount of times it was executed (*Surgery 41 times, Pre_beta 1 time, Post_beta 8 times, Hypotension 111 times, Epidural 141 times and AF 9 times*). Multiple arcs between the same or different activities indicates that the two activities were executed directly one after the other (for example *Hypotension* was directly executed 12 times after *Surgery*). If the activity is the first in the sequence, it will be linked to the starting point (shown as a triangle within a circle) and similarly the last activity in a sequence will be linked to the finishing point (shown as a square within a circle). The analysis was done using Disco [74].

Fig 5 shows that in 88 out of 141 cases, patients will present hypotension after epidural (the epidural activity is followed by the hypotension activity 88 times). This means epidural can cause hypotension, which is consistent with clinical knowledge. Further, it helps to confirm that the absence of a direct dependency between *Epidural* and *Hypotension* in Fig. 4 is most likely because the time we used for reading BP in the BN structure learning might not be immediately after epidural for all of the patients.

Fig. 5 also shows that patients who have epidural either have thoracic *Surgery* or *Pre_beta*, but not both at the same time. This implies that it is not necessary to have *Post_beta* for such epidural patients according to the minimal treatment regime (as explained in section 4.1). Indeed, there are only 8 cases of *Post_beta* in this data set and this might be due to other medical reasons. This explains why there is a direct dependency between *Epidural* and *Post_beta* in Fig 4 because patients who had epidural do not satisfy the two criteria thoracic *Surgery* and *Pre_beta*. Thus, most of the patients who had *Epidural* did not have *Post_beta*.

These two observations show that the work pattern we suggested for epidural patients in Section 4.3.3 is not the case. The model resulting from process mining helps us to confirm the accuracy of the learnt BN structure and also have a better understanding of the underlying reasons for the learnt structure. For example, from the second observation combined with the structure we learnt, we can infer that *Hypotension* affects *Post_beta*, but it is not due to *Epidural*. The combination of BN structure learning and process mining in the work-as-observed largely confirm the safety analysis in the work-as-imagined, except that *Epidural* is not the cause of *Hypotension* in this context (although it might be in others) which is at variance with the safety analysis. This is an example of a clinically meaningful non-alignment between work-as-imagined and the real world – which is exactly what we sought to identify in applying our framework.

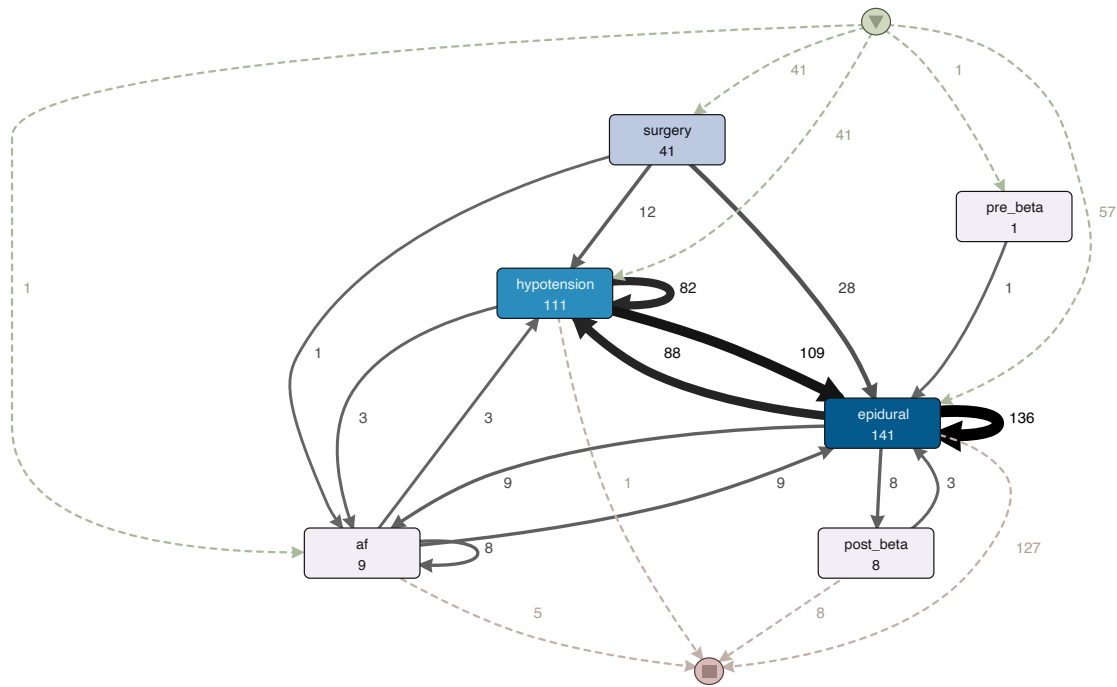

Fig. 5. Process model discovered using Process Mining techniques

The utility of process mining in clarifying and confirming the results of the BN structure learning also suggests that, in the future, we should collect more data for patients who had *Epidural* and also thoracic

*Surgery* and *Pre_beta*, then evaluate the effect of *Epidural* on *Post_beta*. In our framework, this is feedback to refine the data collection aspects of the work-as-observed. This example also illustrates one of the problems of ML in that results can be skewed by bias in the data which we might not be aware of. It also shows the value and utility of combining methods to improve the understanding of ML results.

### 4.3.5. Learning Parameters from Data

Based on the structure learnt in Fig. 4, we used Bayesian estimation to learn the parameters for the network. Once the parameters of the BN have been specified, it allows exploration of the impact of decisions as the context evolves, i.e. probabilistic inference. As specific information about the context is known (e.g. the patient who had *Pre_beta* underwent thoracic *Surgery*), we instantiate the variables corresponding to the context in the network (i.e. *Pre_beta* = 1 *Surgery* = 2), which revises the probability for other variables (e.g. *Post_beta* or *AF*) in the BN to the posterior probability conditioned on the known context. 80% of the data set was used to estimate the parameters, and 20% was used to test them by predicting the development of AF given the values for the remainder of the variables. Table III compares the BN and logistic regression methods for predicting AF. It shows that BN had slightly better prediction accuracy than logistic regression.

TABLE III. PREDICTIVE ACCURACY OF ESTIMATION METHODS

| Methods | Prediction Accuracy | Sensitivity | Specificity |
|---|---|---|---|
| BN | 72% | 6% | 98% |
| LR | 70% | 5% | 96% |

In order to understand the extent to which *Post_beta* affects the development of *AF*, when patients underwent thoracic *Surgery* and had *Pre_beta* (i.e. the effect of Hazard 1) we assessed the posterior probability of developing AF conditioned on *Surgery* = 2, *Pre_beta* = 1 and *Post_beta*. The results are given in Table IV and show that giving BBs after surgery reduces the probability of developing AF from 60% to 49%. In medical terms this is referred to as an 11% absolute risk reduction, or it may be expressed as the number needed to treat of 9 which is good for a medical intervention [75, 76]. This is an important finding as it not only confirms our SHARD analysis, but also is clinically significant. This confirms that giving *Post_beta* to patients who have *Pre_beta* and have had thoracic *Surgery* is beneficial in controlling AF.

Further, in order to assess how significant an influence *Hypotension* has on patients not receiving BB (first row of Table I), we determined the posterior probability of *Post_beta* conditioned on *Surgery* = 2, *Pre_beta* = 1 and *Hypotension*, see Table V. This shows that presenting *Hypotension* decreases the probability of getting *Post_beta* from 45% to 24%. Again, this is an important clinical finding.

TABLE IV. EFFECTS OF POST_BETA ON AF FOR PATIENTS WITH PRE_BETA AND UNDERGOING THORACIC SURGERY

| Development of AF | Post_beta = 0 | Post_beta = 1 |
|---|---|---|
| AF = 0 | 40% | 51% |
| AF = 1 | 60% | 49% |

TABLE V. EFFECTS OF HYPOTENSION ON POST_BETA FOR PATIENTS WITH PRE_BETA AND UNDERGOING THORACIC SURGERY

| Post_beta | Hypotension = 0 | Hypotension = 1 |
|---|---|---|
| Post_beta = 0 | 55% | 76% |
| Post_beta = 1 | 45% | 24% |

4.4. INSTANTIATION OF THE SAFETY CASE

In many engineering domains the use of a safety case is a long-established practice [51]. The safety case is a risk management tool, providing rationale for accepting a system into service and enabling the relevant stakeholders to understand the rationale for using the system and to make informed decisions. Good practice in developing safety cases is to structure them into arguments and evidence, where:

- Argument – the rationale for believing that a system is safe to operate, in its intended context of use;
- Evidence – the available facts or information, e.g. test results, safety analyses, on which the argument rests.

Traditionally, safety cases strongly reflect the safety analysis of the system, or the work-as-imagined, but our framework extends this to include ML analysis of the work-as-observed.

In this paper, we present safety arguments using the Goal Structuring Notation (GSN) [77], which is defined as shown in Fig. 6.

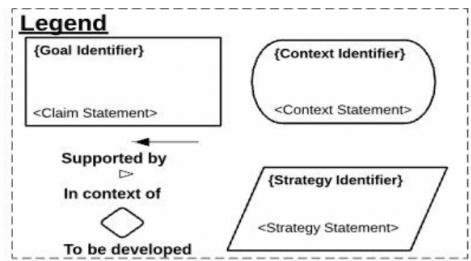

Fig. 6. GSN Notation

In GSN, *goals* represent what we wish to demonstrate. *Goals* can be broken down directly into sub-goals, where the decomposition is clear, or broken down indirectly via *strategies* that explain how the sub-goals relate to the parent goal, where the relationship is not obvious. *Goals* need to be understood in *context*, and *context* is represented using a round-ended rectangle. The use of diamonds below *goals* show that they need further developments. Eventually all *goals* must be supported by evidence. The reader is advised to consult the publicly available GSN standard [78] for a more detailed description of the notation.

A partial argument for control of the risks associated with AF is presented in Fig 7, with explicit traceability to the models included in our framework. The top goal is 'Prevention of AF' – amplified to say control of the risk of AF through use of BBs. The context includes patient characteristics, the hospital setting, assumed to be an ICU, etc.

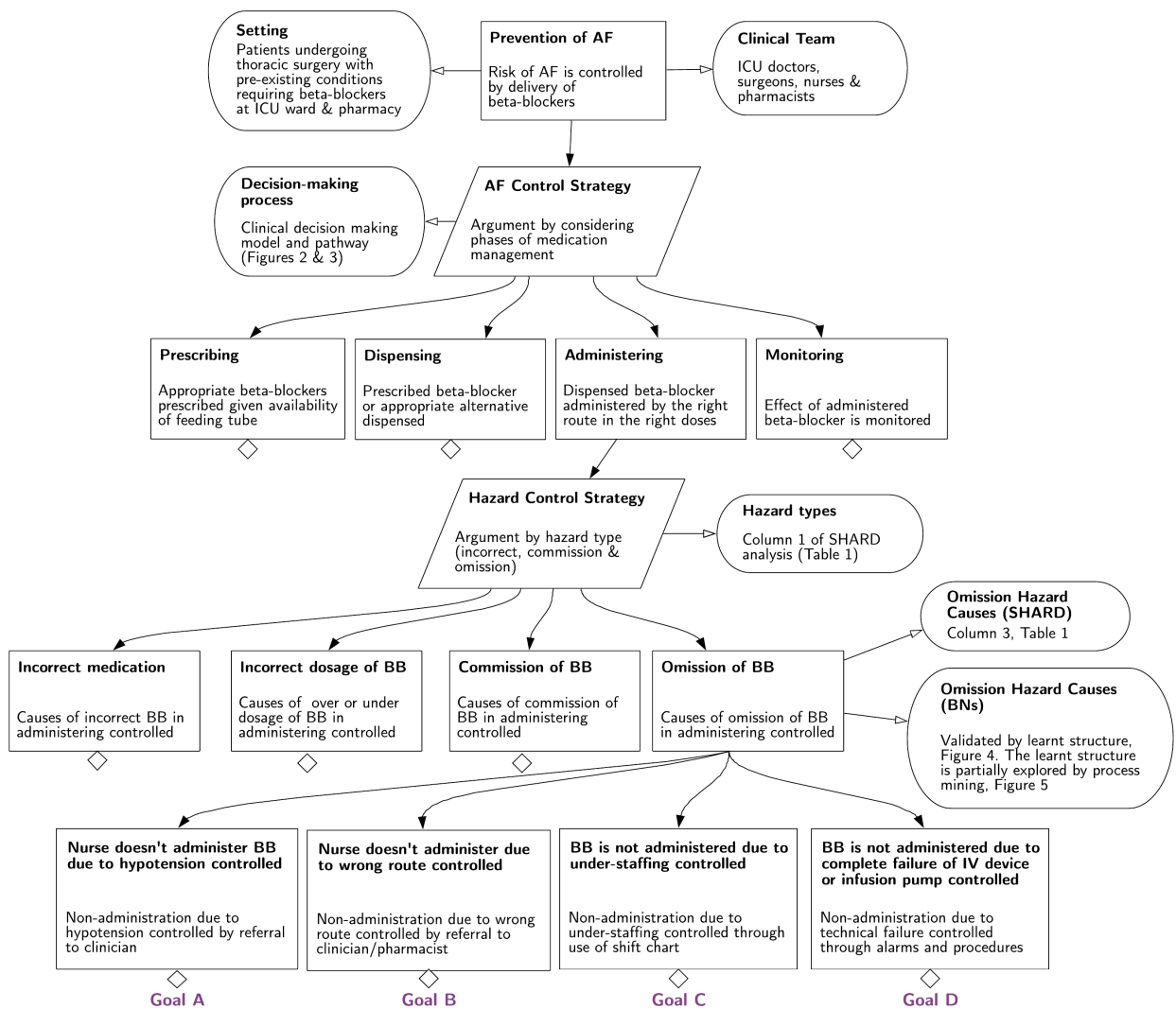

Fig. 7. Safety Argument for Prevention of AF (with Emphasis on Omission of BBs)

The argument is broken down in several layers reflecting the decision-making model (Section 4.2.1) and the safety analysis (Section 4.2.2). The top strategy is a breakdown across the phases of medication management, reflecting the structure of Fig. 3. Below this, the structure is first in terms of hazards (grouping together under and over dosage) and then in terms of controls over the hazard causes. For brevity, Fig. 7 only provides detail on Hazard 1 to illustrate the concept, leaving the rest of the argument undeveloped. The work-as-imagined and work-as-observed come together to provide the context for the Hazard 1, which is 'Omission of BB' goal. The goal needs to be supported taking into account the set of hazard causes that have been 'imagined' and 'observed'; as the combination of the BN structure learning and process mining largely confirm the safety analysis, this correlates strongly with the first row in the SHARD analysis (Table I), but it suggests epidural is not the cause, only hypotension from work-as-observed, thus epidural is not included as a leaf goal in Fig. 7. Further, two of the causes of omission (③ in row 1 of Table I) are from prescribing and dispensing, respectively, so they will not be dealt with under the administering branch.

The four leaf goals, A, B, C and D correspond to the other entries, ①, ②, ④, ⑤ in row 1 of Table I, respectively. Thus, the correspondence between the safety argument, SHARD analysis and the decision-making model can be seen and traced clearly.

Evidence supporting the safety argument (supporting goals A, B, C and D) can be derived from information about the activities in the hospital. Consider goal D – 'BB is not administered due to complete failure of IV device or infusion pump controlled'; in this case data showing the proportion of timely replacements, when devices produce an alarm, would give a 'metric' for the satisfaction of this goal (clearly the closer the proportion is to 100%, the stronger the safety case is). In principle, such information can be derived from hospital records.

The BNs also provide evidence for part of the safety case. For example, as shown in Fig. 4, there is a direct dependency between *Hypotension* and *Post_beta*, which means that the hypothesised cause is real. This gives confidence that the inclusion of goal A – 'Nurse doesn't Administer BB due to hypotension controlled' is necessary. In addition, the BNs also alert us to the level of risk related to Hypotension and show that the presence of *Hypotension* will reduce the chance of getting *Post_beta* by 21%, shown in Table V. This may be a key point to introduce stronger controls (one element of feedback as shown in the framework). After the introduction of new controls, we can use BN to continue to learn from the new data, which should show that the effect of presence of *Hypotension* on getting *Post_beta* is reduced, if the control is effective. However, the BN does not tell us anything explicitly about the controls – such data is not visible in the MIMIC III data set. Further, as MIMIC III does not cover organisational and technical issues, the analysis carried out here does not provide evidence to support goals C and D. We return to these issues in Section 5.

5. DISCUSSION

This paper has proposed a new framework, introduced in Fig. 1, for modern safety management – modern in the sense that it recognises the distinctions between Safety-I and Safety-II, and provides a way of reconciling the two views of safety. We believe that the framework proposed is novel, and has many potential benefits, however it is important to note the status of the work and future developments needed to enable the framework to be widely adopted.

5.1. ROLE AND STATUS OF THE FRAMEWORK

The framework builds on the concepts of Safety-I and Safety-II to provide a practicable means of reducing the gap between the work-as-imagined and work-as-done, by introducing the concept of the work-as-observed. This is intended to be a generic model, reflecting the fact that many systems and situations are now data rich and ML can be used on available data sets to provide an evidence-based way to reduce the gap between the two worlds, as recommended by Hollnagel.

In this paper, we have illustrated the framework on a clinical use case, management of atrial fibrillation in post-operative care, which is both complex and critical, and thus appropriate for assessing the approach. The SHARD analysis gives a systematic way of both identifying hazards and potential causes, drawing on clinical expertise. BN structure learning and process mining are applied to publicly accessible data, namely MIMIC-III, analysing one hazard, failure to administer BBs (omission in the SHARD table). These analyses largely confirm the validity of that part of the safety analysis but identify one area where there is a gap between the world-as-imagined and world-as-observed. Thus, the framework has done its job - and the analyses enable us to identify clinically meaningful reasons for the gap.

The safety methods and ML are mutually supportive. Conducting safety analysis can help us proactively identify the relevant variables to explore in ML and to understand what kind of knowledge we expect to derive, rather than just believing what the ML is telling us (treating it as a 'black box'). The ML is used to validate the safety analysis to show whether or not it is sound and reflecting ground truth rather than being merely hypothetical or subjective or seen as a mere paper-based exercise with little relevance to actual clinical practice.

The framework can also help to combat the potential "*culture of 'paper safety' at the expense of real safety*" which has been a major threat to the validity of safety cases for complex systems in socio-technical environments [79-81]. The safety argument in Fig. 7 shows a systematic way of deriving the safety argument, reflecting both the safety analysis, and the insights from ML. Whilst providing a systematic approach to constructing safety cases can help avoid the problem of 'paper safety' it cannot guarantee that safety cases focus on 'real safety'. However, our use of the safety case which highlights the differences between the work-as-imagined and work-as-observed can inform improvements to safety management - reducing the gap in Hollnagel's terms – and thus go a long way to help with the focus on 'real safety'. As in our case we can't go back to influence the practices in the USA hospitals from which the MIMIC III data was collected it is not possible to show the reduction in the gaps at this stage, but this is part of our longer-term aims.

For brevity, in this paper, we have only addressed one hazard (albeit the most serious one). To confirm the analysis for the other hazards would require different structures to be learnt, based on the other rows in the SHARD table, but the same approach would be used. The data preparation activities would need to be refined for commission and incorrect with clinical input on the effects to be considered. The learning phase of the ML analysis would also follow the same process as described for omission of BBs (hazard 1). The safety argument (Fig. 7) already has the 'placeholders' for addressing these other hazards. To address all the possible causes of the hazards in Table 1 would also require access to other data sources covering technical and organisational factors.

### 5.2. CHALLENGES OF ML

Whilst ML is very powerful, it is also possible to end up with misleading results. Some papers discuss the 'safety' of ML in general [82] and in the medical domain [83]. Here, we first focus on the pragmatic issues that affected the production of the BNs used in this paper, specifically data preparation and process for structure learning, then consider some of the broader issues. In particular we outline approaches to gaining confidence in ML building on the experience with BN structure learning.

A general observation is that data preparation is crucial for ML. It is important to have a good understanding of the data, and of the domain from which the data is drawn, in order to ensure that the selected data is appropriate. If inappropriate, e.g. biased, data is selected for ML then incorrect conclusions could be drawn [84]. This is particularly significant in healthcare as errors in the learnt models might prompt inappropriate interventions, so it is important to refer back to the clinicians and confirm with them what the ML reveals, as has been done here.

Our approach to structure learning for BNs uses the BDeu scoring function, for which we must determine a single hyper-parameter, the equivalent sample size $\alpha$. We found that the learnt structure is quite sensitive to the hyper-parameter $\alpha$. There is a general trend that with the increase of $\alpha$, more arcs are added in the structure, but not necessarily for every increase of $\alpha$. Currently, there is no generally accepted rule for determining the right value of $\alpha$, although there is some ongoing research into rules to set the value of $\alpha$ [85, 86].

Thus, there are challenges in validating the result of ML in a critical situation, including those we just mentioned above. There are several ways of addressing the issue of validating the ML process and the soundness of the learnt models. First, it is possible to combine methods to increase understanding of, and confidence in, the learnt structures as we have done here using BNs and process mining. BN structure learning can show statistical correlations between different factors, no matter whether they occur or not, e.g. the presence of absence of hypotension, and it can also evaluate the effects of one factor on another when the context is known, see Section 4.3.5. In contrast, process mining can only show what has really happened, and cannot explicitly identify what does not happen, as when activities do not happen, there is no event and no timestamp. But it can discover temporal relationships between different activities in the process. Thus, in this case, process mining can help to give further insights into the results of the BN structure learning but is not a substitute for it.

Others have also considered combining ML methods and working on ways of improving the interpretability of learnt models in order to facilitate their scrutiny [87]. We see our combination of BN and process mining as a way of improving the interpretability of the learnt models.

Second, recent work [88] has identified desiderata (or criteria) for ML processes covering data preparation, the learning process including choice of hyper-parameters, as well as model verification. These criteria could give a basis for reasoning about the soundness of the ML process, perhaps using confidence arguments to augment the 'main' safety argument as set out in Fig. 7 [89, 90].

The issue of finding an appropriate and effective strategy for assurance in the ML elements of our framework will be a key consideration in our future work.

### 5.3. LIMITATIONS AND FUTURE WORK

A potential limitation in the case study is that when inferring the development of AF, we used the diagnosis table in MIMIC III to determine which patient has AF after surgery, but because we do not have the medication history information for the patients (maybe due to our limitations in understanding the database), we might include chronic AF in the dataset and not be aware of it. It would be ideal if we could identify and exclude any such patients.

For future work, there are several major steps to realise our overall aims in developing our framework.

First, a complete approach to managing medication error would have to cover the range of factors that can cause and control hazards, i.e. technical and organisational, as well as clinical factors. It would be necessary to obtain data on each factor to enable accurate analysis of the influences on medication error, and undesirable outcomes such as AF. That is, the data would have to cover at least the factors identified in the *Deviation*, *Possible Causes*, and *Detection/Protection* columns in the SHARD analysis (Table I). To provide this data set requires much more extensive preparation, e.g. drawing on hospital administrative data and EPRs, as well as clinical data such as is available in MIMIC III. Thus, an important part of future work is to engage with our clinical partners (Bradford Teaching Hospitals NHS Foundation Trust) to define a data collection mechanism that will enable us to assess all of the non-clinical and clinical factors and to provide actionable recommendations. Some initial work on EPR data has been published [91] but there is more to be done.

Second, to realise the intent of the framework, i.e. to reduce the gap between work-as-imagined and work-as-done, it is necessary to implement the feedback paths shown in Fig. 1. We intend to explore the use of dynamic safety cases as one element of the feedback mechanism. A dynamic safety case [92] has the evidence, and potentially the arguments, updated as the system operates. In the context of our framework, this would mean using ML to continue learning from data in the work-as-observed to update the safety case. To be useful, such updates also need to be associated with criteria for alerting clinicians, e.g. if there is a trend in the data that suggests that planned interventions are ceasing to be effective. However, such work carries considerable risk if it is undertaken incorrectly - so before such activities can be contemplated, it would be important to develop confidence arguments for the use of ML, to be able to demonstrate the safety and effectiveness of the dynamic safety case.

Third, as the framework is intended to be generic it is necessary to apply it more widely, and with different analysis techniques, in order to validate it. In the clinical domain, our intent is to work on sepsis, as this is one of the leading causes of fatalities in hospitals. We also intend to investigate different ML methods. For example, it will be interesting to see whether or not it is possible to apply reinforcement learning as part of the work-as-observed to allow us to learn how the clinicians carry out their work in the real world. Ultimately this might enable the framework to support understanding of the gap between the work-as-imagined and work-as-done in near real-time - however doing this may be very challenging, not least in terms of the willingness of clinicians to accept the recommendations. Here we may be able to learn from the SAM project which is considering clinical acceptability of semi-autonomous systems in ICUs.

6. CONCLUSIONS

This paper has introduced a new framework for assuring medication safety in complex healthcare environments. The framework uses classical safety analysis methods on the normative models to represent the work-as-imagined. It introduces work-as-observed, reflecting the data from the real world, and makes use of ML and other analytical methods to obtain insights into what is happening in the real world, although starting from work-as-imagined. These insights can then be used to inform clinical decision-making. We believe that the combination of safety analysis methods and ML to enable reduction of the gap between the work-as-imagined and work-as-done is a unique approach, especially valuable when systems become more data rich.

We employed a case study focusing on post-operative care following thoracic surgery both to illustrate and to validate the framework. In the case study we were able to identify clinically meaningful results relating to the treatment of thoracic surgery. It also shows the unanticipated mis-alignment between the work-as-imagined and the real world. The case study showed the effectiveness of combining different analysis techniques on the work-as-observed, both to gain further insights into the results of the ML and to give confidence in the results of the analysis. There remain issues in validating the use of ML in critical applications, but we believe that the combination of methods is one important approach to validation.

However, for such a framework to be accepted in the community, it will need the agreement of the regulators. Some regulators are beginning to respond to the issues caused by the introduction of ML into systems. For example, the US Federal Drug Administration (FDA) is developing guidelines on the management of systems incorporating ML, including addressing the ability of systems to change their capabilities through learning in operation [93]. However, this focuses on ML in the system, and does not consider the potential for ML to support safety assurance – it would be interesting to investigate whether or not the FDA's guidelines could be adapted to use of ML for safety assurance.

ACKNOWLEDGMENT

This work draws on two ongoing projects. The core work on medication safety is a project funded by Bradford Teaching Hospitals NHS Foundation Trust. The work on the framework has been carried out in collaboration with the Assuring Autonomy International Programme. The views expressed in this paper are those of the authors and not necessarily those of the NHS, or the Department of Health and Social Care.